\def\year{2020}\relax
\documentclass[letterpaper]{article} 
\usepackage{aaai20}  
\usepackage{times}  
\usepackage{helvet} 
\usepackage{courier}  
\usepackage[hyphens]{url}  
\usepackage{graphicx} 
\usepackage{graphicx}  
\usepackage{amsmath}
\usepackage{amssymb}
\usepackage{algorithm}
\usepackage{algorithmic}
\usepackage{subcaption}
\usepackage{multirow}

\newcommand{\specialcell}[2][c]{\begin{tabular}[#1]{@{}c@{}}#2\end{tabular}}
\usepackage{booktabs}

\newtheorem{Proposition}{Proposition}

\newtheorem{Theorem}{Theorem}
\newtheorem{Corollary}{Corollary}
\newtheorem{Definition}{Definition}
\newtheorem{Problem}{Problem}
\title{Beyond Unfolding: Exact Recovery of Latent Convex Tensor Decomposition \\under Reshuffling }
\author{
Chao Li,\textsuperscript{\rm 1} Mohammad Emtiyaz Khan,\textsuperscript{\rm 1} Zhun Sun,\textsuperscript{\rm 1}\\ 
\Large \textbf{Gang Niu,\textsuperscript{\rm 1} Bo Han,\textsuperscript{\rm 1} Shengli Xie,\textsuperscript{\rm 2,$\ast$} Qibin Zhao\textsuperscript{\rm 1,}\thanks{Corresponding authors. Preprint submitted to AAAI-20.}} \\ 
\textsuperscript{\rm 1}RIKEN Center for Advanced Intelligence Project (AIP), Japan\\ 
\textsuperscript{\rm 2}School of Automation, Guangdong University of Technology, China\\
{\{chao.li, qibin.zhao\}@riken.jp}
}
 
\begin{document}

\maketitle

\begin{abstract}
Exact recovery of tensor decomposition (TD) methods is a desirable property in both unsupervised learning and scientific data analysis. The numerical defects of TD methods, however, limit their practical applications on real-world data. As an alternative, convex tensor decomposition (CTD) was proposed to alleviate these problems, but its exact-recovery property is not properly addressed so far. To this end, we focus on latent convex tensor decomposition (LCTD), a practically widely-used CTD model, and rigorously prove a sufficient condition for its exact-recovery property. Furthermore, we show that such property can be also achieved by a more general model than LCTD. In the new model, we generalize the classic tensor (un-)folding into \emph{reshuffling} operation, a more flexible mapping to relocate the entries of the matrix into a tensor. Armed with the reshuffling operations and exact-recovery property, we explore a totally novel application for (generalized) LCTD, i.e., \emph{image steganography}. Experimental results on synthetic data validate our theory, and results on image steganography show that our method outperforms the state-of-the-art methods.
\end{abstract}

\section{Introduction}
\label{sec:intro}

Tensor decomposition (TD), a multi-linear extension of matrix factorization, has been successfully employed on various applications~\cite{rabusseau2016low,sharan2017orthogonalized,he2017kernelized}.
More importantly, TD methods are also crucial tools for unsupervised discovery of structures hidden behind the data, e.g., localizing the regions of brain from EEG waveforms~\cite{becker2014eeg}, user detection from mobile-communication data, and understanding the kinetic-theory description of materials~\cite{gonzalez2010recent}. 
One of the reasons, behind the success of TD methods in these tasks, is due to the exact-recovery of their solutions, i.e., TD methods are able to achieve a demixing of data in which individual component can tightly correspond with physical interpretation~\cite{williams2018unsupervised}.

The numerical properties of TD methods, however, are not as promising as their exact-recovery property.
The most popular canonical polyadic decomposition (CPD) provides a uniqueness solution up to permutation, but the openness of its solution space and the weak express power are too restrictive for higher-dimensional problems~\cite{comon2009tensor}. Other alternatives, such as (heretical) Tucker decomposition~\cite{cichocki2009nonnegative,songhierarchical}, tensor-train decomposition~\cite{oseledets2011tensor} and tensor-ring decomposition~\cite{zhao2016tensor}, unfortunately do not possess the exact-recovery property, i.e., their components can be arbitrarily rotated without changing the resultant. Block term decomposition (BTD)~\cite{de2008decompositions}, a constrained version of Tucker decomposition, inherits the uniqueness property from CPD but has more flexible decomposition form. However, the model selection for BTD, such as the determination of the multi-linear rank for each block, would be highly challenging. Furthermore, the non-convexity of the aforementioned TD methods generally leads to unstable convergence to global minimum~\cite{kolda2009tensor}.

Convex-optimization-based approaches were proposed to alleviate these unsatisfactory numerical problems~\cite{imaizumi2017tensor,tomioka2010estimation,yu2019tensor}. In these methods, it is not necessary to specify the rank of the decomposition beforehand, and the convexity of the models guarantees both the convergence to the global minimum and their statistical performance. However, there is an important issue that is not properly addressed so far: \emph{Are the convex approaches able to exactly recover the low-rank components like their non-convex counterparts?}

To answer this question, we theoretically prove a sufficient condition for exact-recovery of latent convex tensor decomposition (LCTD), a practically widely-used convex TD method~\cite{yamada2017convex,nimishakavi2018dual} that decomposes a tensor into a mixture of low-rank components. Armed with the notion of incoherence among the components, we rigorously prove that the low-rank components can be exactly recovered when a type of incoherence measure is sufficiently small. Moreover, we show that the exact-recovery property can be owned by a more general class of models than LCTD. In the new model, we introduce the reshuffling operation to replace the conventional tensor (un-)folding used in LCTD, and the new reshuffling operations can give the model the capacity to explore richer low-rank structures. Last, by leveraging the reshuffling operations and the exact-recovery property, we explore a totally novel application of (generalized) LCTD, i.e., image steganography, a classic task in both computer vision and information security. Experimental results not only validate the correctness of our theory, but also demonstrate the model's effectiveness in real-world application.
Supplementary materials are available at: \emph{http://qibinzhao.github.io}.

\subsection{Related Works}

The notion of convex tensor decomposition (CTD) was firstly introduced in~\cite{tomioka2010estimation}, where the decomposition was implemented by minimizing a type of tensor nuclear norm. 
Subsequently, the studies on CTD, especially on latent convex tensor decomposition (LCTD), were continually concerned on both theoretical and practical sides.
LCTD was theoretically proved to achieve tighter upper-bound on the reconstruction performance than its overlapped counterpart~\cite{tomioka2013convex}, and had achieved the state-of-the-art results in many tasks~\cite{guo2017efficient,wang2019latent}.
Works on variants of LCTD are richly proposed recently~\cite{nimishakavi2018dual,wimalawarne2017convex}, but surprisingly the exact-recovery property of LCTD is not paid much attention so far.
More interestingly, in the first paper proposing LCTD~\cite{tomioka2010estimation}, the authors stated that LCTD might not be able to exactly recover the components due to its relation with Tucker model.

For the exact-recovery property of tensor decomposition, the solution of CPD is unique up to permutation under mild condition, while Tucker and TT decompositions are not. 
However, as aforementioned, the numerical problems of these methods, such as the rank determination, somehow lead to difficult implementation in practice.
On the other side, many approaches have focused on restricting the model such that the exact-recovery property is ensured. For example, one way to eliminate the ambiguity in Tucker decomposition is to incorporate additional constraints on the latent factors, e.g., by forcing them to be independent, sparse, or smooth~\cite{cichocki2009nonnegative}. This could work in practice, but in some cases, these constraints might be too strong for the data in hand.
In contrast, we focus on exactly recovering the components by \emph{only} exploiting the low-rank structures of the tensor. 
Additional assumptions such as sparsity, independence are temporarily out of the scope in our work.

The (un-)folding operation used in TD methods builds a connection of the low-rank structures between a matrix and its higher-order form.
In the existing TD methods, this operation is defined by various manipulations~\cite{mu2014square,yu2019tensor}, but one basic principal behind them is to keep more low-rankness of the tensor along the modes.
However, the recent studies on exploiting the low-rank tensor decomposition under general linear transformations~\cite{li2019guaranteed,lu2019low} inspire us that the low-rank structures of a tensor can be explored by more flexible operations than tensor (un-)folding.
In this paper, we therefore generalize the  conventional tensor (un-)folding  into \emph{reshuffling}, which gets rid of the stereotype in the existing definitions of tensor (un-)folding.
In contrast to using arbitrary linear transformation, the proposed reshuffling operations only relocate the entries (without up-sampling) of the data without addition and multiplication, which would result in lower FLOPS in practice.

\section{Latent Convex Tensor Decomposition}

Throughout the paper, we will denote matrices by boldface capital letters, e.g., $\mathbf{X}\in\mathbb{R}^{m\times n}$ is a matrix of size $m\times n$. We will denote tensors by $\mathcal{X} \in\mathbb{R}^{I_1\times I_2\times \cdots\times I_K}$, where $K$ is the \emph{order} of the tensor.
Given data in the form of a tensor $\mathcal{X}$, in this paper we consider low-rank tensor decomposition as the sum of multiple components:
\begin{equation}
\mathcal{X} = \mathcal{A}_1+\mathcal{A}_2+\cdots+\mathcal{A}_N,
\label{eq:tensor_decomposition}
\end{equation}
where $\mathcal{A}_i$ are low-rank components of the same size as $\mathcal{X}$. 
One may argue that the form \eqref{eq:tensor_decomposition} is not low-rank tensor decomposition as it is not in a multiplication of latent factors~\cite{kolda2009tensor}.
Analogous to the singular value decomposition (SVD) of a matrix,
note that both the most popular CPD and Tucker decomposition can be trivially rewritten as~\eqref{eq:tensor_decomposition}, and their ranks determine the number of components~\cite{cichocki2009nonnegative}.

Latent convex tensor decomposition (LCTD) can be also formulated as similar as \eqref{eq:tensor_decomposition}. LCTD decomposes a tensor into the sum of components but incorporating the (un-)folding operations on each component, i.e., 
\begin{equation}
    \begin{split}
        \mathcal{X}=\Phi_1(\mathbf{A}_1)+\Phi_2(\mathbf{A}_2)+\cdots+\Phi_K(\mathbf{A}_K),
    \end{split}\label{eq:LCTD}
\end{equation}
where the matrices $\mathbf{A}_i$ are assumed to be low-rank, and the operations $\Phi_i(\cdot)$ are also called folding or tensorization in literature. 
In contrast to CPD and Tucker decomposition, LCTD explore the low-rank structures of the matricized form of each component, and its algorithm is based on convex optimization.

The exact-recovery property of LCTD, however, is still a kind of open problem. The consistency of $\mathbf{A}_i$ in \eqref{eq:LCTD} was discussed in~\cite{tomioka2013convex}, but the reconstruction bound does not tend to be zero when decreasing the strength of the noise. It implies that there is still a theoretical gap between consistency and identiability, a.k.a., the exact-recovery of components. 

In this paper, we will try to fill this gap by imposing the notion of incoherence as the condition for the exact-recovery.
We find that the different (un-)folding operations on components could bring us incoherent low-rank structures, which has been proved as an important characteristic in compressed sensing and matrix completion~\cite{candes2009exact}.
Meanwhile, we also find by the numerical experiments that the unbalance between the number of rows and columns of the unfolded tensor is probably one key reason that restricts the capacity for exactly recovering the components. To this end, we generalize
the conventional tensor (un-)folding into reshuffling, which is able to result in more balanced and incoherent model.

\section{Reshuffled Tensor Decomposition}
In this section, we introduce a generalization of LCTD, where low-rank matrices are mapped to tensors by the reshuffling operations.
To be self-contained of the paper, we also briefly derive an algorithm with stable convergence.

\subsection{Formulation}
We consider the following model similar to \eqref{eq:LCTD}:
\begin{equation}
    \mathcal{X}=R_1(\mathbf{A}_1)+R_2(\mathbf{A}_2)+\cdots+R_N(\mathbf{A}_N),\label{eq:genModel}
\end{equation}
where $\mathbf{A}_i\in\mathbb{R}^{m_i\times{}n_i},i\in{}[N]$ denotes the latent components. However, in contrast to \eqref{eq:LCTD}, we arbitrarily choose the number of components, and replace the folding operations $\Phi_i(\cdot)$ with a more general and flexible type for operations $R_i:\mathbb{R}^{m_i\times{}n_i}\rightarrow\mathbb{R}^{I_1\times{}I_2\times{}\cdots\times{}I_K}$, which are a group of linear operators called \emph{reshuffling} and defined as follows:
\begin{Definition}[Reshuffling]
The reshuffling operation, denoted by $R$, is defined as a mapping that maps a matrix $\mathbf{A}\in\mathbb{R}^{m\times{}n}$ into a real tensor $\mathcal{A} := R(\mathbf{A})$ of size ${I_1\times{}I_2\times\cdots\times{}I_K}$, such that the number of elements in $\mathcal{A}$ is equal to the number of elements in $\mathbf{A}$, i.e., $mn = I_1I_2\ldots I_K$, and every entry of $\mathbf{A}$ correspond to one and only one entry in $\mathcal{A}$.
\label{def:reshuffling}
\end{Definition}
The main idea behind our model is to employ a variety of reshuffling operations for every component, i.e., for each component $\mathbf{A}_i$,
we reshuffle it using a distinct operator $R_i$ to get a tensor $\mathcal{A}_i:=R_i(\mathbf{A}_i)$. Meanwhile, we assume that the rank of $\mathbf{A}_i$ to be small, and would like to be able to capture a variety of low-rank structures within the tensor. 

We build a convex optimization problem for recovering the components $\mathbf{A}_i$ from Eq. \eqref{eq:genModel}. Assuming that the reshuffling operations $R_i$ are known for each component, $\mathbf{A}_i,\,i\in{}[N]$ are recovered by minimizing the following optimization problem:
\begin{equation}
    \begin{split}
        &\min_{\mathbf{A}_1,\ldots,\mathbf{A}_N}\sum_{i=1}^N\Vert\mathbf{A}_i\Vert_*,\, \mathrm{s.t.},\,\mathcal{X}=\sum_{i=1}^N{}R_i(\mathbf{A}_i)
    \end{split},\label{eq:Reshuffled-TD}
\end{equation}
where $\Vert\cdot\Vert_*$ denotes the matrix nuclear norm, which equals the sum of the singular values of matrix. It has been proved that the matrix nuclear norm is the convex envelope of matrix rank~\cite{fazel2001rank}. Hence, we can roughly say that solving the problem \eqref{eq:Reshuffled-TD} is equivalent to looking for the most low-rank latent components from the observed tensor $\mathcal{X}$.

\textbf{Remark 1:}
Similarly to the conventional folding operation, reshuffling maps a matrix into a tensor, and is a linear and reversible operator.
However, reshuffling can operate more flexibly and handle the matrices of arbitrary size.
\emph{The conventional tensor (un-)folding can be therefore obtained as a special case of reshuffling.}
The flexibility of reshuffling could enable recovery of low-rank structures that were previously unrecoverable by using folding operations. 

\textbf{Remark 2:} The reshuffling operations for particular data could be difficult to find. In some applications, such as stenography (shown in Experiment Section), the operations are known beforehand. In general, one could design them to exploit some specific characteristics of the data. For example, the unfolding operation exploits the physical meaning associated with the modes to convert the tensor into a matrix. Reshuffling operations could be designed in a similar fashion to exploit other types of structural information about the tensor. 
In this paper, we focus on discussing the exact-recovery (conditions) of the method and assume the reshuffling operations to be known in advance.

\subsection{Algorithm}

Below, we derive an algorithm called reshuffled tensor decomposition (Reshuffled-TD) to solve \eqref{eq:Reshuffled-TD}.

Due to the existence of the equality constraints, we apply the augmented Lagrangian method for solving \eqref{eq:Reshuffled-TD}, of which the Lagrangian function is given by
\begin{equation}
    \begin{split}
        &L\left(\mathbf{A}_1,\ldots,\mathbf{A}_N,\mathcal{Y},\kappa\right)=\sum_{i=1}^N\Vert\mathbf{A}_i\Vert_* \\
        &+\left<\mathcal{Y},\mathcal{X}-\sum_{i=1}^N{}R_i(\mathbf{A}_i)\right>+\frac{\kappa}{2}\left\Vert\mathcal{X}-\sum_{i=1}^N{}R_i(\mathbf{A}_i)\right\Vert_F^2,
    \end{split}
\end{equation}
where the tensor $\mathcal{Y}$ denotes the Lagrangian dual and $\kappa>0$ is a positive scalar. In the algorithm, we sequentially update $\mathbf{A}_i,i\in{}[N]$, $\mathcal{Y}$ and $\kappa$ in each iteration. As the key step of the algorithm, we update $\mathbf{A}_i$
by minimizing the following sub-problem:
\begin{equation}
    \begin{split}
        \mathbf{A}_i^{+}=&\arg\min_{\mathbf{A}_i}\frac{1}{\kappa}\Vert\mathbf{A}_i\Vert_* \\
        &+\frac{1}{2}\left\Vert\mathbf{A}_i-R_i^\star\left(\mathcal{X}-\sum_{j\neq{}i}R_j(\mathbf{A}_j)+\kappa^{-1}\mathcal{Y}\right)\right\Vert_F^2,
    \end{split}\label{eq:UpdateA}
\end{equation}
where $R_i^\star$ denotes the adjoint operator of $R_i$. 
In contrast to the renowned Alternating Direction Method of Multipliers (ADMM) methods, we update the scalar $\kappa$ in each iteration by multiplying a constant $\rho$ larger than 1 (for example $\rho=1.01$), i.e., $\kappa^+=\rho\kappa$, and the work in~\cite{lin2010augmented} shows that such trick could efficiently accelerate the convergence compared to ADMM.
The complete procedure of Reshuffled-TD is given in Alg.~\ref{alg:Reshuffled_TD}, where $D_\alpha(\,\cdot\,)$ denotes the soft-thresholding operation on the singular values~\cite{cai2010singular}.
\begin{algorithm}[t!]
\caption{Reshuffled-TD}
\label{alg:Reshuffled_TD}
\begin{algorithmic}
\STATE \textbf{Initialize}: The observation $\mathcal{X}$, $\mathcal{Y}=sgn(\mathcal{X})$, $\mathbf{A}_i=R_i^\star(\mathcal{X})/N,\,\forall{}i$, and $\rho>1,\kappa_0>0$
\STATE \textbf{Iteration until convergence:}
\STATE //STEP 1: update every latent component $\mathbf{A}_i,\,i\in{}[N]$
\STATE\quad \textbf{for} $i=1,\ldots,N$ \textbf{do}
\STATE\quad \quad{}Update $\mathbf{A}_i$ by using
\begin{equation*}
    \mathbf{A}_i\leftarrow{}D_{\kappa^{-1}}
    \left({}R_i^\star\left(\mathcal{X}-
    \sum_{j\neq{}i}R_j(\mathbf{A}_j)+\kappa^{-1}\mathcal{Y}
    \right)\right)
    \label{eq:STH}
\end{equation*}
\STATE\quad \textbf{end for}
\STATE //STEP 2: update the Lagrangian dual $\mathcal{Y}$
\STATE\quad $\mathcal{Y}\leftarrow\mathcal{Y}+\kappa\left(\mathcal{X}-\sum_{i=1}^N{}R_i(\mathbf{A}_i)\right)$
\STATE //STEP 3: update the scalar $\kappa$
\STATE\quad$\kappa\leftarrow\rho\kappa$
\STATE \textbf{Output}: $(\mathbf{A}_i,\ldots,\mathbf{A}_N)$
\end{algorithmic}
\end{algorithm}

Next,
we show that the algorithm results in convergence to the optimal point of \eqref{eq:Reshuffled-TD}. For the brevity of the proof, we only consider the case when $N=2$ in the theorem, and from the experimental results we also find the stable convergence even in the case with more components ($N\geq{}3$).

\begin{Theorem}
Using Reshuffled-TD and assuming the number of the components $N=2$, if the sequence $\{\kappa^{(k)}\}$ is non-decreasing, and $\sum_{k=1}^\infty{}\kappa^{(k),-1}=+\infty$, then $(\mathbf{A}_1^{(k)},\mathbf{A}_2^{(k)})$ converges to an optimal solution of (\ref{eq:Reshuffled-TD}).\label{Theorem:convergence}
\end{Theorem}
The proof is given in the supplemental material.

It is worth noting that the convergence of Reshuffled-TD does not ensure whether the solution is equal to the ``true'' components that give rise to the observed tensor. 
Therefore, for the guarantee of the exact-recovery, we need to take more structural assumptions of the ``true'' components into account.

\section{Exact Recovery with Reshuffled-TD}
\label{sec:theoretical}
In this section, we derive and prove the exact-recovery conditions when using the Reshuffled-TD method. We start with a formal statement of the problem. 
\begin{Problem}[Conditions for Exact Recovery]
Given a tensor $\mathcal{X}$, suppose there exist low-rank matrices $\mathbf{A}_i^*$ with rank $k_i$ such that $\mathcal{X} = \sum_{i=1}^N{}R_i(\mathbf{A}_i^*)$. Under what conditions on $\mathbf{A}_i^*, R_i$ and $k_i$, the estimated $\hat{\mathbf{A}}_i$, obtained by using Reshuffled-TD, will be equal to $\mathbf{A}_i^*$ for all $i$?
\label{ProblemSetting}
\end{Problem}

Our solution for the problem is stated in Theorem \ref{thm:exact_recovery}. The main result relies on an \emph{incoherence} (defined in Definition \ref{def:incoherence}) which measures the change in the rank of a component $\mathbf{A}_i^*$ when the operation $R_i$ is replaced by any other operator $R_j$, i.e. from $R_i^\star(\mathcal{A}_i^*)$ to $R_j^\star(\mathcal{A}_i^*)$ where $\mathcal{A}_i^*:=R_i(\mathbf{A}_i^*)$ and $R^\star$ denotes the adjoint of $R$. 
To be able to measure this change, we first need to define a low-rank manifold over tensor for a given $R_i$ (see Definition \ref{def:P}), and a neighborhood in this manifold. For the latter, we will show a type of the tangent space in this manifold (see Proposition \ref{def:T}). We start with the formal definition of the manifold. 

\begin{Definition}[Tensor manifold under reshuffling]
Given a reshuffling operation $R_i$, the following set of tensors $\mathcal{Y}$ such that the rank of the matrix $\mathbf{Y}=R_i^\star(\mathcal{Y})$ is equal to $k_i$
\begin{equation}
\mathbb{P}_i:=\left\{\mathcal{Y}\in{}\mathbb{R}^{I_1\times{}\cdots\times{}I_K}\vert{}
rank(\mathbf{Y})=k_i\right\}.
\label{P}
\end{equation}
defines a smooth manifold~\cite{hosseini2017tangent}.
\label{def:P}
\end{Definition}
We now define a neighborhood in $\mathbb{P}_i$ using a type of tangent space. 
In the derivation, the tangent space around a tensor $\mathcal{A} \in \mathbb{P}_i$ is obtained by the truncated singular-value decomposition of $\mathbf{A}=R_i^\star(\mathcal{A})$ where $\mathbf{A}\in\mathbb{R}^{m_i\times{}n_i}$ is a rank-$k_i$ matrix. Truncated SVD of $\mathbf{A}$ with first $k_i$ leading singular values is given by $\mathbf{U}\mathbf{\Lambda}\mathbf{V}^\top$, where $\mathbf{U}$ and $\mathbf{V}$ are matrices of size $m_i \times k_i$ and $n_i\times k_i$ respectively, and $\mathbf{\Lambda}$ is a diagonal matrix that contains the $k_i$ singular values as its diagonal. By considering all possible real matrices of size $m_i\times k_i$ and $n_i\times k_i$, the tangent space of $\mathbb{P}_i$ at the point $\mathcal{A}=R_i(\mathbf{A})$ is given by the following proposition.
\begin{Proposition}[Tangent Space]
Given a rank-$k_i$ matrix $\mathbf{A}\in{}\mathbb{R}^{m_i\times{}n_i}$, which generates a tensor $\mathcal{A}$ by $R_i$, i.e., $\mathcal{A}=R_i(\mathbf{A})$, the tangent space of the manifold $\mathbb{P}_i$ at the given tensor $\mathcal{A}$ is formalized as the following,
\begin{align}
    &\mathbb{T}_i(\mathbf{A}) := \left\{\mathcal{Y} \vert{}\right.\\
    &\left.R_i^\star(\mathcal{Y}) = \mathbf{U}\bar{\mathbf{V}}^\top +\bar{\mathbf{U}}\mathbf{V}^\top, \bar{\mathbf{U}} \in\mathbb{R}^{m_i\times{}k_i}, \bar{\mathbf{V}}\in\mathbb{R}^{n_i\times{}k_i}\right\}. \nonumber
\end{align}
\label{def:T}
\end{Proposition}

The proof of the proposition can be trivially achieved from Eq. (3.2) in~\cite{chandrasekaran2011rank}. The tangent space $\mathbb{T}_i(\mathbf{A})$ gives us an approximation of the manifold in a neighborhood of $\mathcal{A}$. Due to the relationship $\mathcal{A}=R_i(\mathbf{A})$, $\mathbb{T}_i(\mathbf{A})$ can be used to analyze how perturbation influences the rank of $\mathbf{A}$. This is captured in the following incoherence measure, which we define next.
\begin{Definition}[Reshuffled-low-rank incoherence]
Consider the tangent space $\mathbb{T}_i(\mathbf{A})$ in the manifold $\mathbb{P}_i$ of (true) rank $k_i$.
Given a different operation $R_j$, we look at all the tensors $\mathcal{Y} \in \mathbb{T}_i(\mathbf{A})$, and find the maximum spectral norm $\|R_j^\star{}(\mathcal{Y})\|_2$ while $\|R_i^\star(\mathcal{Y})\|_2 < 1$ for the $i$'th operator.
The incoherence of a tensor $\mathcal{A}=R_i(\mathbf{A})$ is then defined to be the maximum spectral norm obtained for all operations $R_j\ne R_i$. Formally,
\begin{equation}
\begin{split}
\mu_i\left(\mathbf{A}\right)&:=\max_{j\ne i} 
\max_{ \begin{array}{c}
\mathcal{Y}\in\mathbb{T}_i(\mathbf{A}),\\\left\Vert{}R_i^\star\left(\mathcal{Y}\right)\right\Vert_2\leq{}1
\end{array}}
\left\Vert{}R_j^\star{}\left(\mathcal{Y}\right)\right\Vert_2.\\
\end{split}\label{eq:incoherence}
\end{equation}
\label{def:incoherence}
\end{Definition}
The above incoherence measure captures the change in the rank when the operation is changed from $R_i$ to any other $R_j$. 
This is due to a relationship between the spectral norm and the rank. The spectral norm is the dual of the nuclear norm which is a convex surrogate of the matrix rank. Roughly speaking, when the spectral norm under $R_i$ is constrained, a small spectral norm obtained under $R_j$ would imply a large change in the rank of the reshuffled matrices. Therefore, a small value of the incoherence measurement would imply an increase in the rank when the true operator is replaced by a different one.

Our main result is to show that bounding the incoherence measurement ensures exact recovery.
\begin{Theorem}[Exact-Recovery Condition]
The estimated $\hat{\mathbf{A}}_i$, obtained by Reshuffled-TD, are equal to the true $\mathbf{A}_i^*$ for all $i$, when
\begin{equation}
\max_{i=1,\ldots,N}\mu_i(\mathbf{A}_i^*)<\frac{1}{3N-2},
\end{equation}
where $N$ denotes the number of the components.
\label{thm:exact_recovery}
\end{Theorem}
The above condition states that if incoherence measurements are small enough, then exact-recovery is possible. Roughly, this implies that, for exact recovery, the rank of components must increase drastically whenever we switch its corresponding reshuffling operation to other else. 
From the geometric view, it implies that the tangent spaces $\mathbb{T}_i(\mathbf{A}_i^*),\,\forall{}i$ need to be well ``separated'' to each other.

Using Theorem \ref{thm:exact_recovery} and imposing more constraints on the reshuffling operation $R_i$, we can get a more intuitive condition for the exact recovery:
\begin{Corollary}
Assume that $\mathcal{X}=\sum_{i=1}^N{}R_i(\mathbf{A}_i^*)$ is a $K$th-order tensor with the size $I\times{}\ldots\times{}I$, and the the reshuffling operations $R_i:\mathbb{R}^{n\times{}n}\rightarrow\mathbb{R}^{I\times{}\ldots\times{}I}$ for each component. In addition, suppose that (a) the rank of $\mathbf{A}_i^*$ equals $r$; (b) it is full-rank for all matrices $R_j^\star\left(R_i(\mathbf{A}_i^*)\right),\forall{}j\neq{}i$; (c) For each matrix $R_j^\star\left(R_i(\mathbf{A}_i^*)\right),\,\forall{}i,j$, its non-zero singular values are equal to each other. Then $(\mathbf{A}_1,\ldots,\mathbf{A}_N)=(\mathbf{A}_1^*,\ldots,\mathbf{A}_N^*)$ is the unique solution of \eqref{eq:Reshuffled-TD} if $n>\left(3N-2\right)^2r$.\label{CorollarySquare}
\end{Corollary}
It implies from Corollary \ref{CorollarySquare} that, the lower bound of the size $n$ will linearly changed with the rank $r$, but quadratically changed with the number of components $N$ for the exact recovery. Although assumptions in Corollary \ref{CorollarySquare} is quite strict, the result still reveals an intuitive fact that the true components can be more likely exactly recovered by our method if the data size $n$ is large enough.

Our proof builds upon some of the techniques used in~\cite{chandrasekaran2011rank} to prove similar results for a type of matrix decomposition. Our proof extends these techniques to the tensor with multiple reshuffled-low-rank structures. 
Compared to the theoretical studies in \cite{tomioka2013convex}, we focus on the conditions for the exact recovery while they mainly analyze the statistical performance influenced by the perturbation like Gaussian noise. Although the Theorem 2 and 3 in~\cite{tomioka2013convex} shows that the upper bound for the sum of the reconstruction error of components tends to be tighter with decreasing the strength of the perturbation, the upper bound is \emph{not} guaranteed to go to zero even though the strength of the perturbation goes to zero. However, we rigorously prove that the decomposition can exactly recover the latent components, and explicitly give the incoherence condition on exact recovery for the first time.

\section{Experimental Results}
\label{sec:exp}

In the experiments, we specify the reshuffling operations by uniformly random permutation for simplicity. In practice, the reshuffling operations can be determined by more practical rules or prior knowledge.

\subsection{Validation of Exact-Recovery Conditions}

\begin{figure*}
        \begin{subfigure}[b]{0.25\textwidth}
                \centering
                \includegraphics[width=.85\linewidth]{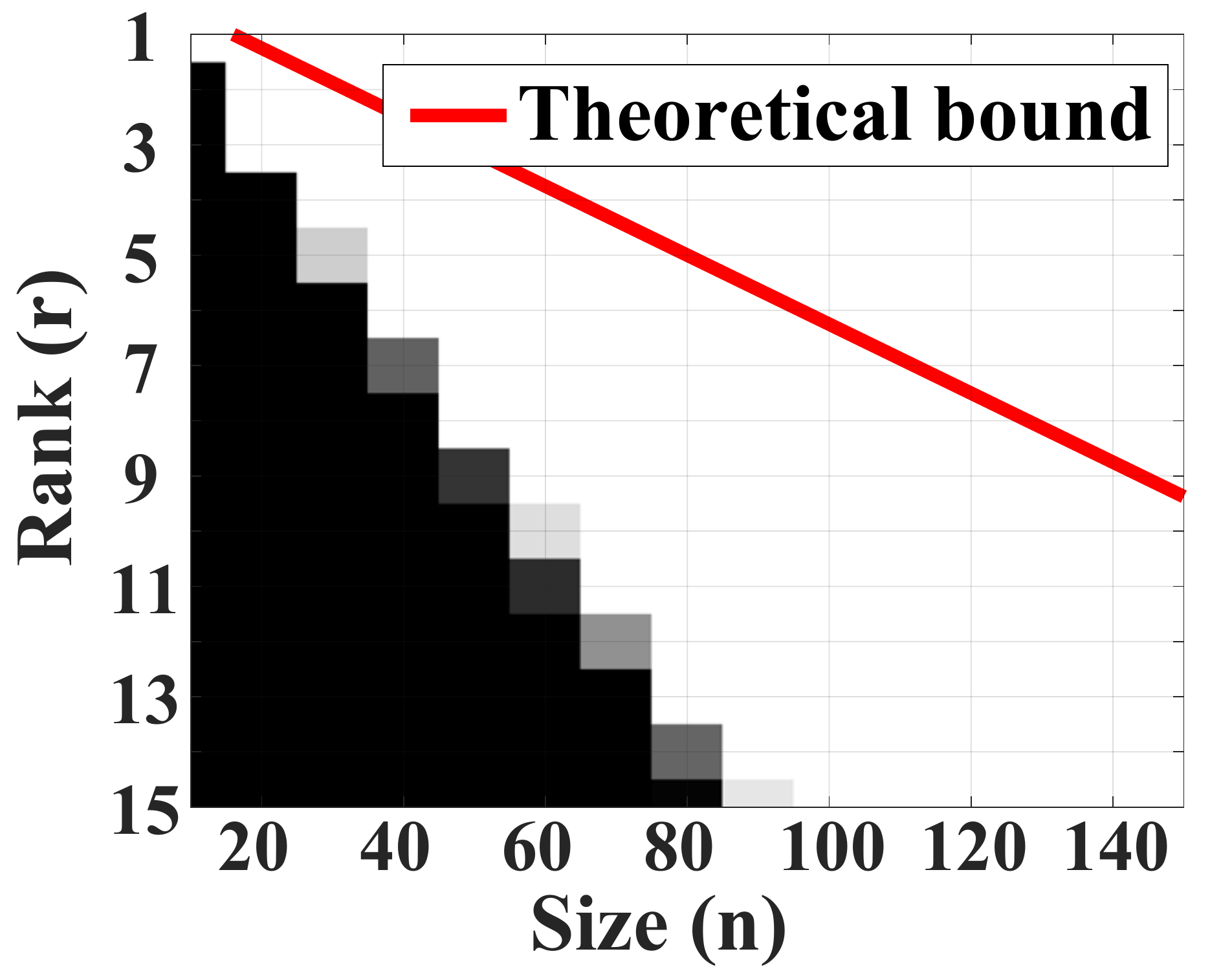}
                \caption{$N=2$}\label{fig1:a}
                \label{fig:gull}
        \end{subfigure}%
        \begin{subfigure}[b]{0.25\textwidth}
                \centering
                \includegraphics[width=.85\linewidth]{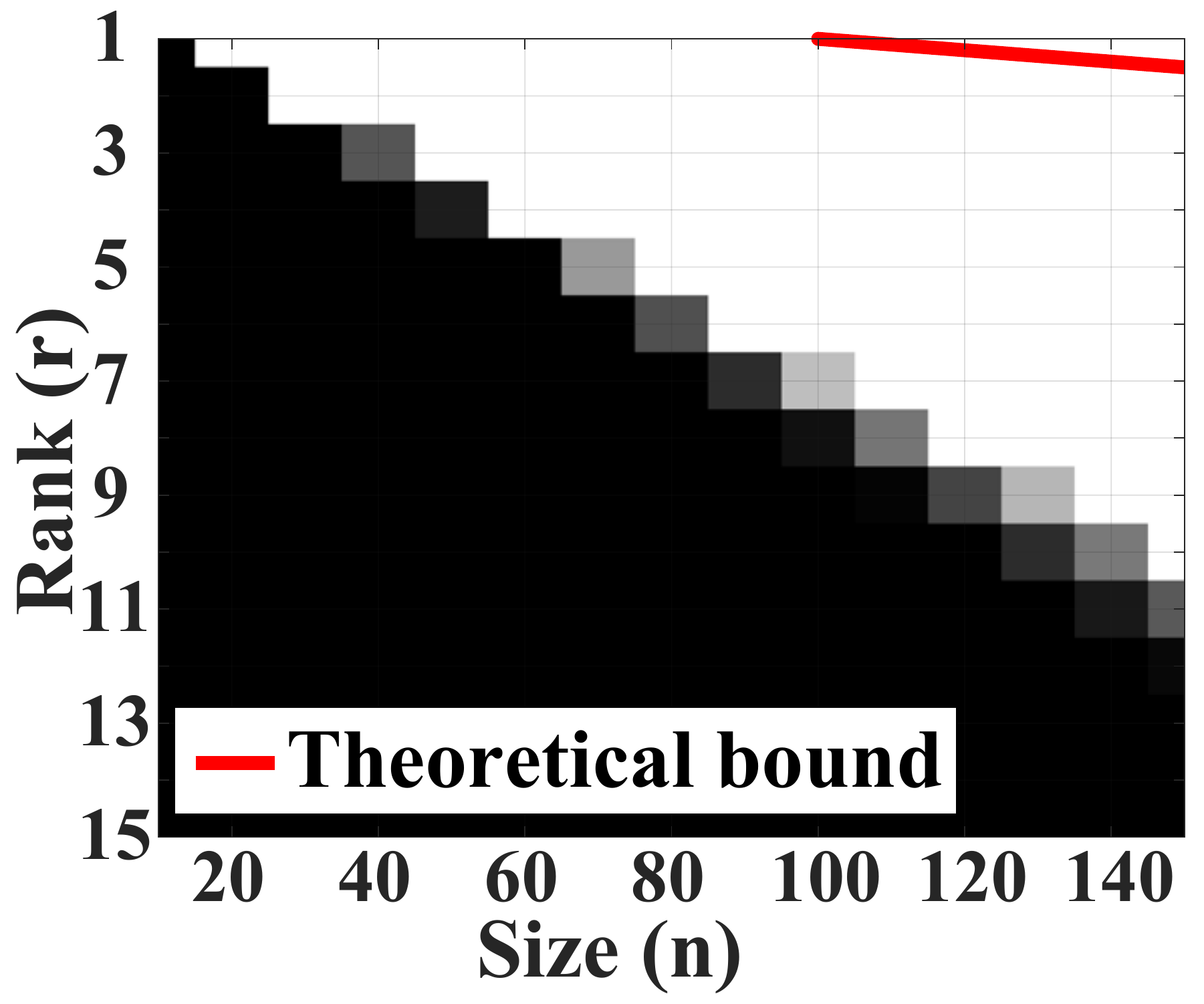}
                 \caption{$N=4$}\label{fig1:c}
                \label{fig:gull2}
        \end{subfigure}%
        \begin{subfigure}[b]{0.25\textwidth}
                \centering
                \includegraphics[width=.85\linewidth]{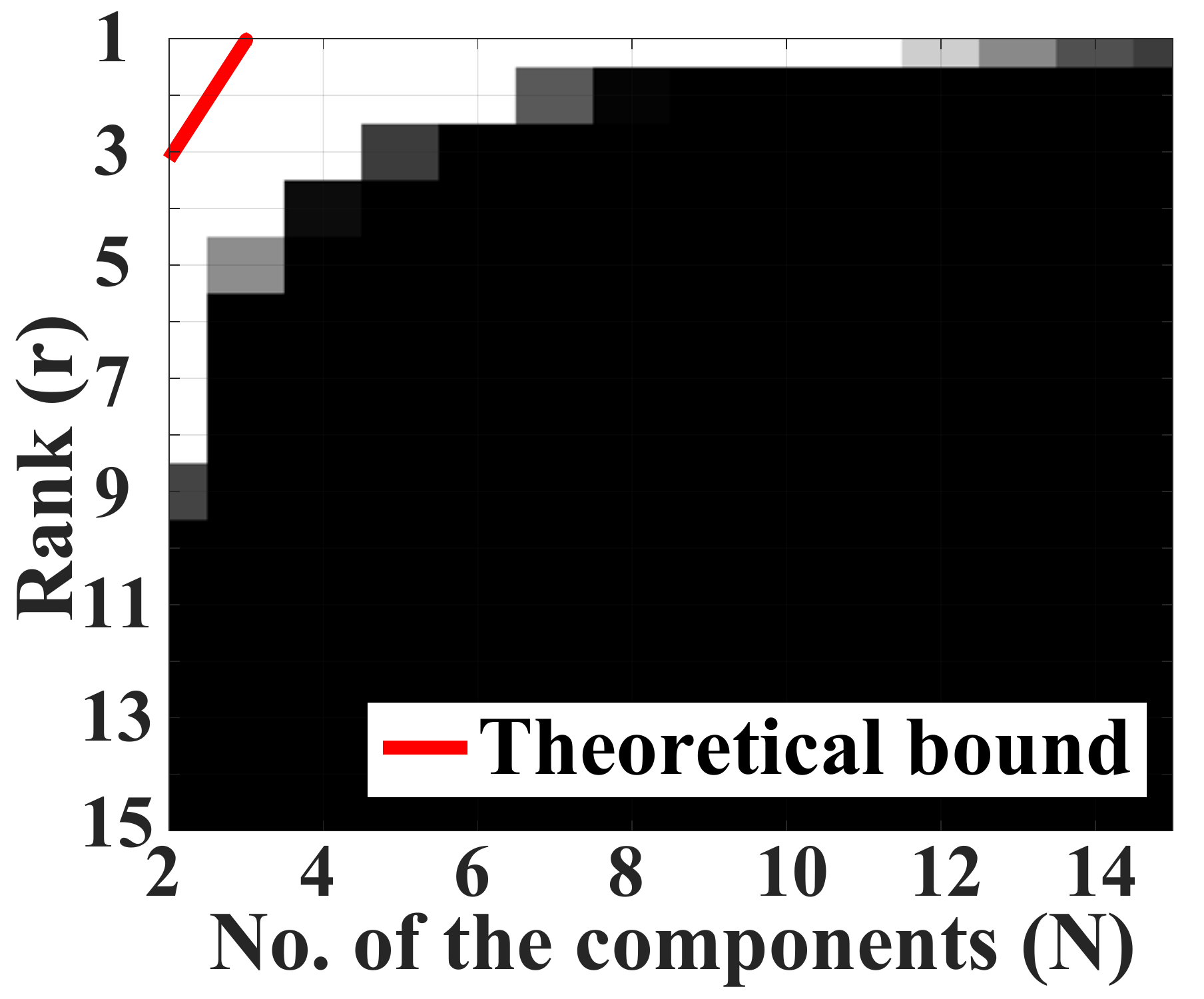}
                \caption{$n=50$}\label{fig1:d}
                \label{fig:tiger}
        \end{subfigure}%
        \begin{subfigure}[b]{0.25\textwidth}
                \centering
                \includegraphics[width=.85\linewidth]{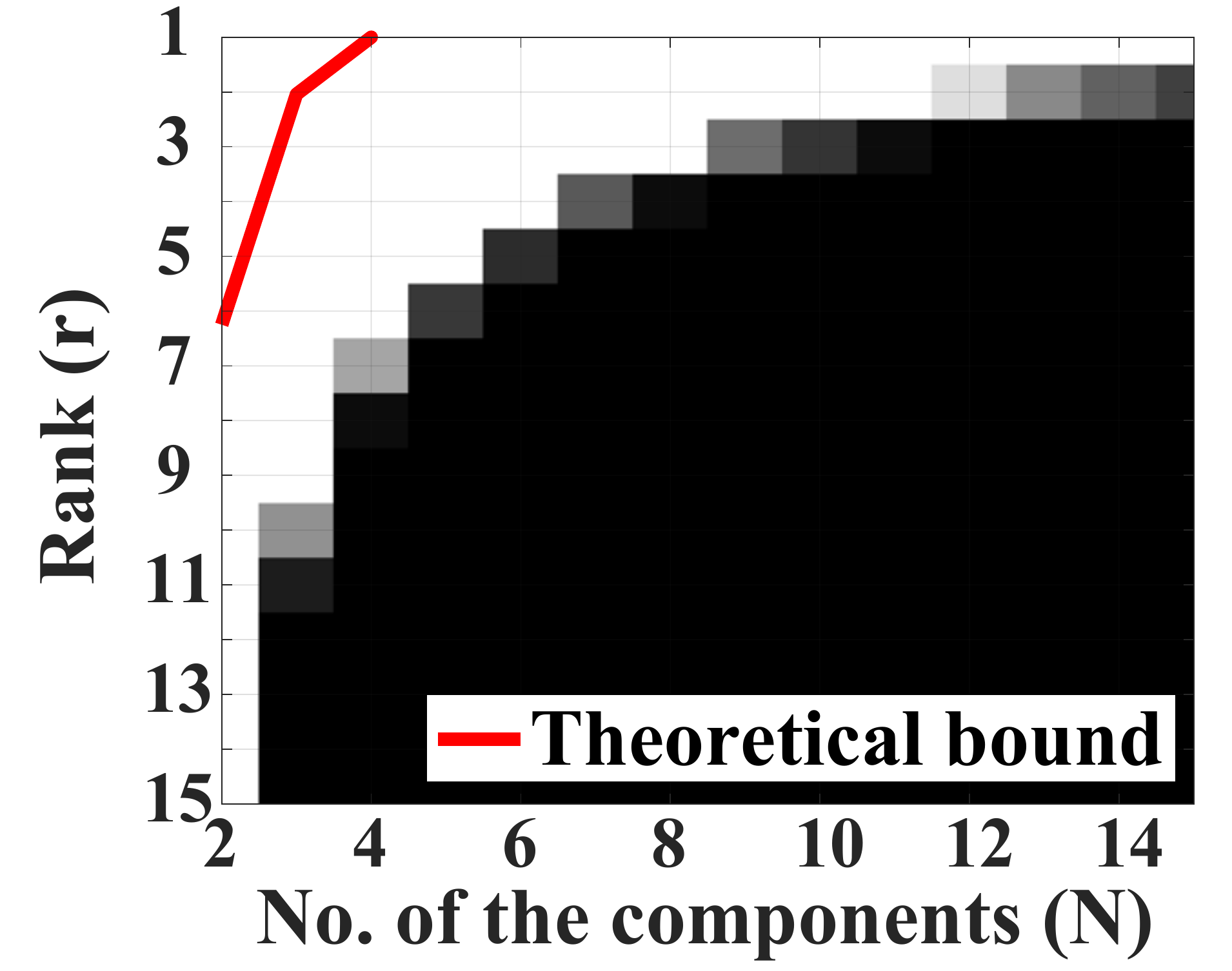}
                \caption{$n=100$}\label{fig1:f}
                \label{fig:mouse}
        \end{subfigure}
        \caption{Results on synthetic data to validate our exact-recovery results. We vary different experiment parameters, such as rank $r$, size $n$ and  number of the components $N$. In each plot, the darker areas denote the worse reconstruction ($tSIR\leq{}15$dB) while the white areas denote a good recovery ($tSIR\geq{}25$dB). The gray boundary shows the phase transition, while the red line denotes the phase transition predicted by our theoretical bound derived in Corollary 1.}
        \label{Fig_1}
\end{figure*}

We firstly perform an experiment using synthetic data to validate the theoretical results in the paper. We generate data by using $N$ square matrices $\mathbf{A}_i^*\in\mathbb{R}^{n\times{}n}$, $i\in{}[N]$. Each $\mathbf{A}_i^*$ is generated by multiplying two random semi-orthonormal matrices with rank $r$, i.e., $\mathbf{A}_i^*=\mathbf{U}_i\mathbf{V}_i^\top$ in which $\mathbf{U}_i,\mathbf{V}_i\in\mathbb{R}^{n\times{}r}$ denote the random semi-orthonormal matrices. 

We measure the performance using the total signal-to-interference ratio (tSIR) defined as follows: 
\begin{align}
    \textrm{tSIR} :=\sum_{i=1}^N\left\Vert\mathbf{A}_i^*\right\Vert_F^2/\sum_{i=1}^N\left\Vert\mathbf{A}_i^*-\hat{\mathbf{A}}_i\right\Vert_F^2
\end{align}
Fig.~\ref{Fig_1} shows the phase transition of Reshuffled-TD with different parameters, such as the rank $r$ and size $n$ of the matrices $\mathbf{A}_i^*$ and the number of components $N$. In each plot, the white blocks indicate $tSIR\geq{}25dB$ which implies very good recovery, and the black blocks indicate $tSIR\leq{}15dB$ which implies no recovery. The gray area corresponds the results in between and indicates the phase transition from exact recovery to partial or no recovery. This can be compared with the theoretical bound given in Corollary \ref{CorollarySquare} which is shown with the red line. From Corollary 1 we can find, for a fix $N$, the relationship between $n$ and $r$ is linear, and, when $n$ is fixed, the relationship between $N$ and $r$ is quadratic. This matches the relationship shown from experimental results. Our bound is a bit conservative, but correctly captures a major chunk of the area where exact recovery is possible. 

\begin{figure}[t!]
    \centering
    \begin{subfigure}[t]{0.48\columnwidth}
        \centering
        \includegraphics[height=1.2in]{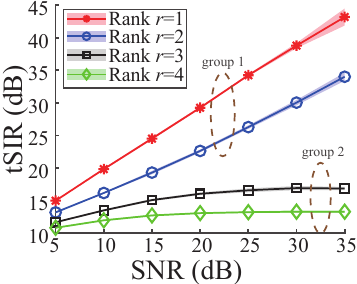}
        \caption{Gaussian Noise}
    \end{subfigure}%
    ~ 
    \begin{subfigure}[t]{0.48\columnwidth}
        \centering
        \includegraphics[height=1.2in]{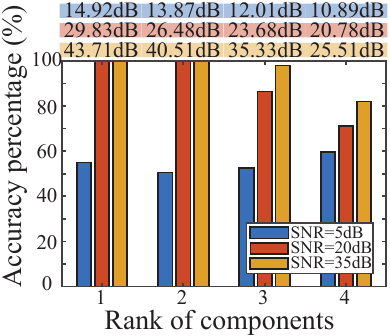}
        \caption{Components' number determination}
    \end{subfigure}
    \caption{Illustration on robustness of Reshuffled-TD on non-ideal experimental conditions. Sub-figure (a) shows the tSIR performance of the method which is influenced by Gaussian noise. In the sub-figure, group 1 corresponds the rank which satisfies the exact-recovery condition, while the rank in group 2 is not. Sub-figure (b) shows the estimation accuracy of the number of components, and the values shown above the bars are the corresponding tSIR performance of the proposed method.}\label{fig:SNRandMissing}
\end{figure}

\textbf{Noise Robustness of Reshuffled-TD.} We impose the Gaussian noise to evaluate the impact on the performance of Reshuffled-TD. Specifically, we fix the size of the components $n=100$, the number of the components $N=10$ and set the rank of each component by $r=1,\ldots,4$. Then, we add the zero-mean i.i.d. Gaussian noise to the data, and the variance of the noise is controlled by the signal to noise ratio (SNR). Fig. \ref{fig:SNRandMissing} (a) illustrate the performance of Reshuffled-TD when $5dB\leq{}SNR\leq{}35dB$.

As shown in \ref{fig:SNRandMissing}~(a), four performance curves are split into $2$ groups.
we know from Fig.~\ref{Fig_1}~(d) that group 1 corresponds the rank which satisfies the exact-recovery condition (in the white area), while group 2 corresponds the rank whose values do not satisfy the conditions (in the back area). Hence, the two groups have different trend with the variety of SNR. In addition, we can find that, in group 1, tSIR is larger than $20$dB when $SNR\geq{}20$dB. It implies that our method works smoothly with high SNR. Because we do not consider the noise in our model (it can be seen from the equality constraint), the performance of our method becomes worse when SNR is low.

\textbf{Robustness on the Number of the Components}. We consider the case that the proposed method does not exactly know how many components are contained in the observation. To simulate this situation, we randomly remove some components from original 10 components, and the removal probability for each component satisfies the Bernoulli distribution (the mean value equals $0.5$). As to the proposed method, we still assume that all $10$ components are contained in the data.  To estimate the number of components by Reshuffled-TD, we compare the norm of the recovered components with a threshold (we choose $\eta=0.1$ for numerical consideration). Fig.~\ref{fig:SNRandMissing} (b) illustrates the estimation accuracy. Besides the accuracy, the corresponding tSNR performance is also shown above the bar-plot in the figure.

As shown in Fig.~\ref{fig:SNRandMissing}~(b), the proposed method is able to reconstruct the contained components with high performance even if the the true number of the components is less than expectation. With high SNR value (SNR$\geq{}20$), the estimation accuracy of the number of the components achieves $100\%$, and the accuracy decreases when choosing a large rank. The high accuracy of our method is due to the fact that the exact-recovery conditions can be still theoretically satisfied as long as we assign an incoherent reshuffling operation $R_i$, even if the norms of some components equal zero.

\subsection{Image Steganography using Reshuffled-TD}
Steganography is about concealing a secret message within an ordinary message and then extracting it at its destination. In this experiment, we will use Reshuffled-TD for image steganography, i.e., to conceal a ``secret'' image in an ordinary ``cover'' image.

Image steganography is a classic problem for both computer vision and information security. In the existing methods, the most popular one is the least-significant-bits (LSB) method, which uses the least significant bits of the cover to hide the most significant bits of the image. In addition, the similar idea is also extended to transform domains like Fourier and wavelet~\cite{holub2014universal}.

Some recent approaches have used deep neural networks to hide and recover images~\cite{baluja2017hiding}, but these methods require lots of training data, and they are generally sensitive to the images not present in the training data. The computational requirement is also heavy. In contrast, our method is much simpler. It does not require any training, and therefore does not have any such sensitivity issues. 

We tried various ways to make the problem challenging for the methods. We try to conceal a full-size RGB image ($8\times{}3$ bits per pixel) into a grayscale image (8 bits per pixel). Meanwhile, we choose different types of images for steganography, e.g., natural, cartoon and fingerprint. 

{
\begin{table}[t]
\small
  \centering
  \caption{Performance comparison of image steganography. In the experiment, we use SIR (dB) to quantify the distortion of both cover(C) and secret(S) images, where larger value of SIR indicates better performance. For each row, the SIR value will be highlighted if it outperforms other methods under a given parameter.}
  \label{table:performance}
  \resizebox{0.95\columnwidth}{!}{\begin{tabular}{*{12}{c}}
    \toprule
    \multirow{3}[2]{*}{\textbf{Datasets}} & 
        \multicolumn{2}{c}{\textbf{LSB}} & 
        \multicolumn{2}{c}{\textbf{DWT}} &
        \multirow{3}[2]{*}{\textbf{DPS}} &
        \multirow{3}[2]{*}{\textbf{Ours}} \\
    \cmidrule(lr){2-3}\cmidrule(lr){4-5}
     & 1 bit/ & 2 bits/ & 2 bits/ & 3 bits/  & & \\
     & chn & chn & chn & chn  & & \\
    \midrule
    \multirow{2}[0]{*}{\specialcell{\textbf{DTD(C)} \\ \textbf{CART.(S)}}} & 
    26.70 & 9.66 & 
    25.17 & 23.45  & 
    --- & 
    20.40  \\
     & 6.92 & 14.42 & 
    12.90 & 17.81  & 
    14.04 & 
    21.64 \\
    \midrule
    \multirow{2}[0]{*}{\specialcell{\textbf{DTD(C)} \\ \textbf{DTD(S)}}} &   
    23.77 & 7.53 & 
    22.81 & 19.40  & 
    --- & 
    \textbf{23.69} \\
     & 3.38 & 7.84 & 
    5.27 & 9.05  & 
    3.43 & 
    \textbf{11.36} \\
    \midrule
    \multirow{2}[0]{*}{\specialcell{\textbf{DTD(C)} \\ \textbf{FIVEK(S)}}} &   
    24.05 & 7.76 & 
    24.70 & 22.27  & 
    --- & 
    \textbf{23.36} \\
     & 1.12 & 6.00 & 
    4.69 & 8.48 & 
    8.97 & 
    \textbf{10.87} \\
    \midrule
    \multirow{2}[0]{*}{\specialcell{\textbf{FIVEK(C)} \\ \textbf{FIVEK(S)}}} &   
    23.02 & 6.56 & 
    21.54 & 18.57  & 
    --- & 
    21.86 \\
     & 3.37 & 7.52 & 
    5.48 & 8.74  & 
    8.96 & 
    6.67 \\
    \midrule
    \multirow{2}[0]{*}{\specialcell{\textbf{FVC(C)} \\ \textbf{FIVEK(S)}}} &   
    18.19 & 3.27 & 
    24.47 & 19.95  & 
    --- & 
    \textbf{20.25} \\
     & 3.32 & 6.42 & 
    4.84 & 8.30  & 
    8.90 & 
    \textbf{12.80} \\
    \midrule
    \multirow{2}[0]{*}{\specialcell{\textbf{LIVE(C)} \\ \textbf{FIVEK(S)}}} &   
    24.50 & 7.66 & 
    24.49 & 20.93  & 
    --- & 
    \textbf{24.71} \\
     & 4.08 & 7.58 & 
    5.32 & 9.46  & 
    9.50 & 
    \textbf{11.49} \\
    \bottomrule
  \end{tabular}}
\end{table}}

\begin{figure*}
    \centering
        \includegraphics[width=0.9\textwidth]{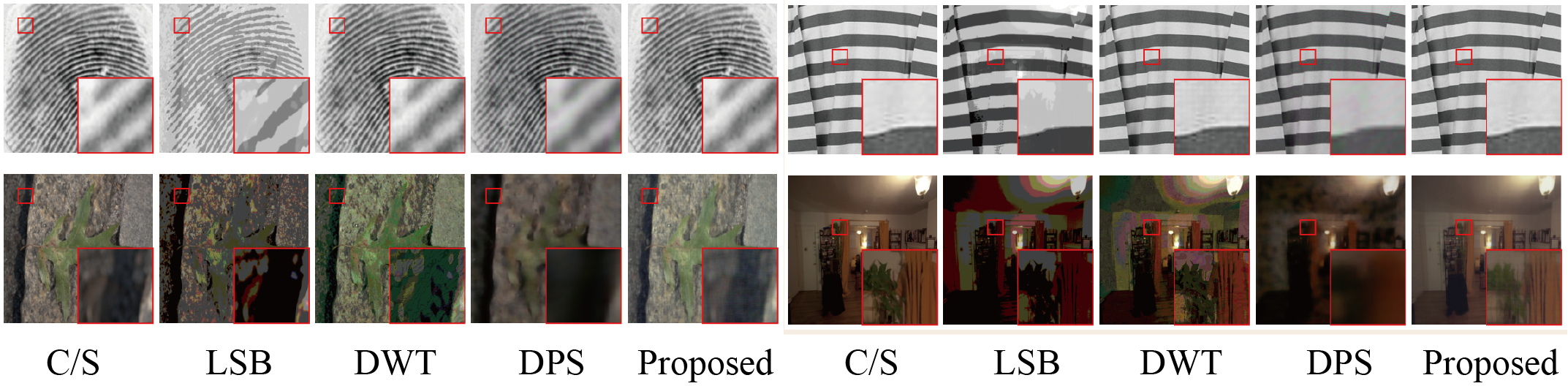}
    \caption{An example to illustrate the performance of image steganography by different methods. In the figure, the first  row shows the original cover images (the first column) and the container images generated by different methods; The second row shows the original secret images (the first column) and its recovery by different methods. 
    }
    \label{fig:example}
\end{figure*}

The datasets we used in the experiment include texture (DTD), natural (LIVE and FIVEK~\cite{fivek}), cartoon (CART.~\cite{royer2017xgan}) and fingerprint (FVC~\cite{maltoni2009handbook}) datasets.
For different datasets, we unify the shape of all images to $2000\times{}2000$, and convert the image to grayscale when the cover image is colored. 

A sketch of our Reshuffled-TD method is shown in supplemental materials. During the concealing phase, 
we consider each channel of the secret image as one component, and they are randomly reshuffled.
Then, we added the reshuffled ``components'' to the cover image to obtain the a grayscale ``container'' image (a.k.a. observation). The difference between the container and cover images will tend to zero as we decrease the strength of the secret components by multiplying a scalar. Therefore, we expect that the secret image can be hidden well if we choose a appropriate value of this ``strength'' scalar. In the recovery phase, we use the reshuffling operations as  key, and recover the RGB components of the secret image by Reshuffled-TD.

Experimental results are shown in Table \ref{table:performance} as measured by the signal to interference ratio (SIR). A higher value of SIR indicate better performance. The experiment is conducted on 10 randomly chosen image pairs. We compare to three existing methods: (a) the LSB method (b) the discrete wavelet transform based method (DWT), and deep stego (DPS)~\cite{baluja2017hiding}. Because DPS converts the grayscale cover image into a RGB image as the container, we just show the SIR for secret image in the table. 

As shown in Table \ref{table:performance}, Reshuffled-TD significantly outperforms all the state-of-the-art methods in the experiment.
For example, in FVC+FIVEK dataset, Reshuffled-TD achieves 20.25dB on the cover images and 12.80dB on the secret images. With the similar SIR on the cover image, LSB only achieves 3.32dB. Meanwhile, DWT and DPS  achieve 8.30 and 8.90dB, respectively. As the worst performance, the FIVEK+FIVEK dataset shows an exception in the experiment. This is because FIVEK is a dataset of natural images, which contains full of detail information. Hence, there is less room in the cover images to hide additional information. Fig. \ref{fig:example} shows two examples of reconstructed images obtained in the experiment. More examples for visual comparison are shown in the supplementary material.

\section{Conclusion}
By leveraging the (generalized) latent convex tensor decomposition, a.k.a. Reshuffled-TD, we rigorously proved that the low-rank components an be exactly recovered when the incoherence is sufficiently upper-bounded. In addition, we applied the generalized model to a totally novel task, i.e. image steganography.
Experimental results on various real-world datasets demonstrate that Reshuffled-TD outperforms both the classic state-of-the-arts but also deep-learning-based methods.

As potential works in the future, first we consider to take the low-tensor-rank assumption into the model instead of the current multi-linear rank.
Second, it would be an interesting topic to seek for the ``optimal'' reshuffling operations if there exist training data. If such operations are learnable, it implies hat we might find lower-rank representation for the data, such that the well-developed low-rank-based methods can be employed on the transformed data.

\section{Acknowledgments}
This work was partially supported by JSPS KAKENHI (Grant No. 17K00326) and NSFC (Grant No. 61673124).

\bibliographystyle{aaai}
\bibliography{aaaibib}
\end{document}